\documentclass{article}
\PassOptionsToPackage{dvipsnames,table}{xcolor}
\usepackage{arxiv}
\usepackage{graphicx}
\usepackage[numbers]{natbib}
\usepackage{doi}
\usepackage[utf8]{inputenc} 
\usepackage[T1]{fontenc}    
\usepackage{url}            
\usepackage{booktabs}       
\usepackage{amsfonts}       
\usepackage{nicefrac}       
\usepackage{microtype}      
\usepackage{amsmath}
\usepackage{amssymb}
\usepackage{bbm}
\usepackage{wasysym}
\usepackage{pifont}
\usepackage{threeparttable}
\usepackage{subcaption}
\usepackage{array} 
\usepackage{makecell}
\usepackage{algpseudocode} 
\usepackage{multirow}
\usepackage{listings}
\usepackage{enumitem}
\usepackage{hyperref}
\usepackage{todonotes}
\usepackage{comment}
\usepackage[capitalise, noabbrev]{cleveref}
\usepackage{lscape}
\usepackage{tabularray}
\usepackage{CJK}

\newcommand{\eg}{\textit{e.g.}}

\newcommand{\ie}{\emph{i.e.}}

\usepackage{amsfonts}
\usepackage{amsmath}
\usepackage{multirow}
\usepackage{amssymb}
\usepackage{pifont}
\usepackage{xcolor}
\usepackage{colortbl}
\usepackage{makecell}
\usepackage{wrapfig}

\definecolor{tabfirst}{rgb}{1, 0.7, 0.7}
\definecolor{tabsecond}{rgb}{1, 0.85, 0.7}
\definecolor{tabthird}{rgb}{1, 1, 0.7}

\let\titleold\title
\renewcommand{\title}[1]{\titleold{#1}\newcommand{\thetitle}{#1}}

\title{Language and Geometry Grounded Sparse Voxel Representations for Holistic Scene Understanding} 

\author{Guile Wu\textsuperscript{1}, David Huang\textsuperscript{1,2,*}, Bingbing Liu\textsuperscript{1}, and Dongfeng Bai\textsuperscript{1}\\
\textsuperscript{1}Huawei Noah's Ark Lab \quad \textsuperscript{2}University of Toronto \\
{\tt\small guile.wu@outlook.com, dawae.huang@mail.utoronto.ca, \{liu.bingbing, baidongfeng\}@huawei.com}
}

\renewcommand{\headeright}{Technical Report}

\begin{document}

\maketitle
\begin{center}
    \centering
    \captionsetup{type=figure}
    \includegraphics[width=0.9\linewidth]{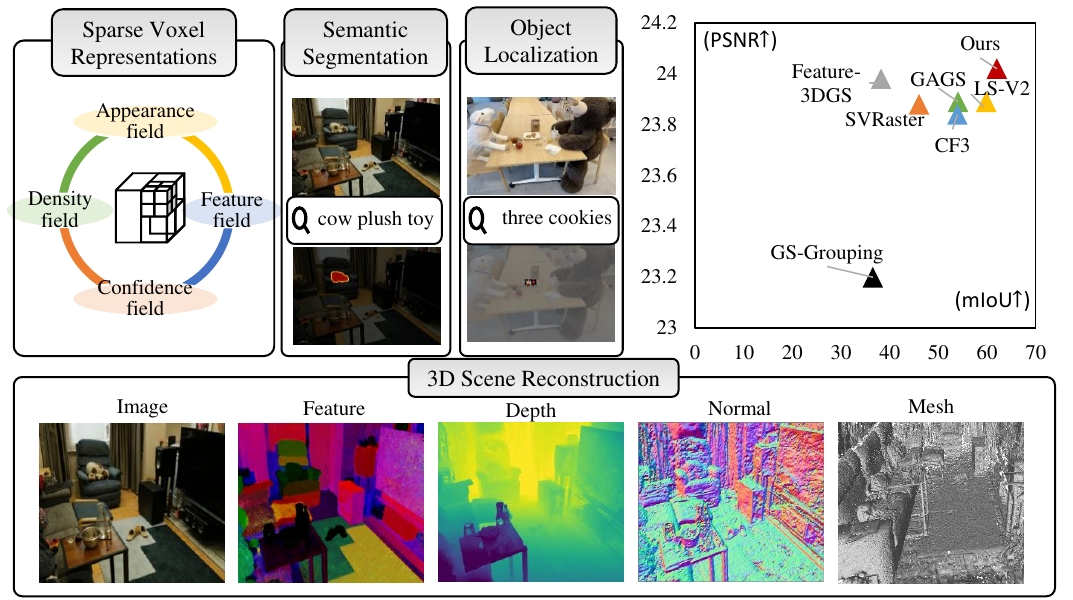}
    \captionof{figure}{
        An illustration of our approach to holistic scene understanding.
        Our approach leverages language and geometry grounded sparse voxel representations to comprehensively model the appearance, semantics, and geometry of the 3D scene in a unified framework, achieving better overall performance compared with the state-of-the-arts.
        }
    \label{fig_illustration}
\end{center}%

\begingroup
\renewcommand\thefootnote{*}
\footnotetext{David Huang contributed to this work during an internship at Huawei Canada.}
\endgroup

\thispagestyle{fancy}

\begin{abstract}
Existing 3D open-vocabulary scene understanding methods mostly emphasize distilling language features from 2D foundation models into 3D feature fields, but largely overlook the synergy among scene appearance, semantics, and geometry.
As a result, scene understanding often deviates from the underlying geometric structure of scenes and becomes decoupled from the reconstruction process.
In this work, we propose a novel approach that leverages language and geometry grounded sparse voxel representations to comprehensively model appearance, semantics, and geometry within a unified framework.
Specifically, we use 3D sparse voxels as primitives and employ an appearance field, a density field, a feature field, and a confidence field to holistically represent a 3D scene.
To promote synergy among the appearance, density, and feature fields, we construct a feature modulation module and distill language features from a 2D foundation model into our 3D scene model.
In addition, we integrate geometric distillation into feature field distillation to transfer geometric knowledge from a geometry foundation model to our 3D scene representations via depth correlation regularization and pattern consistency regularization.
These components work together to synergistically model the appearance, semantics, and geometry of the 3D scene within a unified framework.
Extensive experiments demonstrate that our approach achieves superior overall performance compared with state-of-the-art methods in holistic scene understanding and reconstruction.
\end{abstract}
    
\section{Introduction}
\label{sec:introduction}

3D scene reconstruction has made significant progress in recent years~\cite{kerbl20233d,sun2025sparse,mildenhall2021nerf,barron2022mip}.
Methods such as Neural Radiance Fields (NeRF)~\cite{mildenhall2021nerf} and 3D Gaussian Splatting (3DGS)~\cite{kerbl20233d} have demonstrated excellent performance in high-fidelity scene reconstruction and have been widely used in various applications, such as robotics, augmented reality, and autonomous driving.
However, most 3D scene reconstruction methods focus on modeling scene appearance and geometry but overlook semantic feature learning, limiting their ability to produce high-level, semantically informative scene representations.

On the other hand, some studies~\cite{qin2024langsplat,qu2024goi,peng2026gags,li2025langsplatv2} have proposed distilling language features from 2D foundation models, such as CLIP~\cite{radford2021learning}, into feature fields of 3D scene models.
A popular strategy is to pre-train a 3DGS model and optimize additional feature fields via feature distillation~\cite{qin2024langsplat,qu2024goi,li2025langsplatv2}.
While these methods have shown promising performance for open-vocabulary scene understanding, they primarily emphasize feature field distillation and largely overlook scene geometry modeling.
Moreover, these methods decouple 3D scene understanding from the reconstruction process, which deviates from the underlying geometric structure of scenes and results in sub-optimal scene understanding and reconstruction.
Although a few methods~\cite{shi2024language,ye2024gaussian,kerr2023lerf} have explored a one-stage paradigm for scene understanding, they usually perform worse than their two-stage counterparts and have not yet fully exploited the synergy among appearance, semantics, and geometry for holistic scene understanding and reconstruction.

In this work, we propose a novel approach named \textbf{LangSVR}, which learns language and geometry grounded sparse voxel representations to comprehensively model scene appearance, semantics, and geometry in a unified framework for holistic scene understanding and reconstruction.
Specifically, we use sparse voxels~\cite{sun2025sparse} as 3D primitives and employ an appearance field, a density field, a feature field, and a confidence field to model a 3D scene.
Through a differentiable rasterizer, our model can render various maps, such as RGB images, feature maps, depth maps, normal maps, and confidence maps, to comprehensively represent a 3D scene.
To promote the synergy among appearance, density, and feature fields, we construct a feature modulation module and distill language features from a 2D foundation model into our 3D scene model.
Meanwhile, to capture the underlying geometric structure of scenes, we distill geometric knowledge from a geometry foundation model into our 3D scene representations via depth correlation regularization and pattern consistency regularization.
Moreover, to enhance multi-view consistency, we employ confidence maps to filter out noisy representations.
These components work together to synergistically model the appearance, semantics, and geometry of the 3D scene, facilitating holistic 3D scene understanding and reconstruction (see Fig.~\ref{fig_illustration}).
Our experimental results demonstrate that our approach achieves superior overall performance compared with state-of-the-art methods in various holistic scene understanding and reconstruction tasks, such as 3D semantic segmentation, 3D object localization, and novel view synthesis.
In summary, our \textbf{contributions} are:
\begin{itemize}
  \item {We propose language and geometry grounded sparse voxel representations to facilitate the synergy among scene appearance, semantics, and geometry for holistic 3D scene understanding and reconstruction.}
  \item {We integrate geometric distillation into feature field distillation to transfer geometric knowledge from a geometry foundation model to 3D scene representations.}
  \item {We conduct extensive experiments on different datasets and demonstrate the superiority of our approach over state-of-the-art methods in holistic scene understanding and reconstruction.}
\end{itemize}

\section{Related Work}
\label{sec:related_work}

\subsection{Language Feature Field Distillation}
Feature field distillation focuses on transferring feature knowledge from 2D foundation models into 3D models (such as NeRF and 3DGS)~\cite{kerr2023lerf,qin2024langsplat,li2025langsplatv2,peng2026gags,zuo2025fmgs,lee2025cf3,wang2025ag2aussian,zhou2024feature,peng20253d}.
LERF~\cite{kerr2023lerf} grounds feature embeddings from off-the-shelf models, such as CLIP~\cite{radford2021learning}, into NeRF for open-vocabulary language queries in 3D space.
LangSplat~\cite{qin2024langsplat} optimizes a language field in 3DGS for 3D scene understanding by distilling multi-granularity language features from CLIP into a collection of 3D Gaussians.
LEGaussians~\cite{shi2024language} and LangSplatV2~\cite{li2025langsplatv2} propose dedicated quantization strategies for more efficient feature embeddings in 3DGS.
Although these methods have shown promising performance, they mostly overlook the synergy among appearance, semantics, and geometry in 3D models.
Unlike existing works, our work proposes novel language and geometry grounded sparse voxel representations to synergistically model the appearance, semantics, and geometry of the 3D scene, which facilitates holistic scene understanding and reconstruction.

\subsection{3D Scene Reconstruction}
3D scene reconstruction is a long-standing topic in computer vision and has made remarkable progress in recent years.
Generally, 3D scene reconstruction can be categorized into implicit and explicit modeling.
Implicit modeling methods, such as NeRF~\cite{mildenhall2021nerf}, represent a 3D scene with a continuous radiance field~\cite{mildenhall2021nerf,barron2022mip,barron2021mip}.
Although these methods achieve compelling performance, they require computationally intensive dense sampling and complex neural networks, resulting in lower training and rendering efficiency.
In contrast, explicit modeling methods, \eg, 3DGS~\cite{kerbl20233d} and sparse voxel rasterization (SVRaster)~\cite{sun2025sparse}, model a 3D scene with explicit primitives (such as 3D Gaussians and voxels) and differentiable rasterizers~\cite{kerbl20233d,yu2024mip,sun2025sparse}.
Among them, 3DGS has achieved great success due to its high efficiency and has been widely used in various areas~\cite{kocabas2024hugs,fan2024lightgaussian,wu2025armgs,qin2024langsplat,ren2024unigaussian}.
More recently, SVRaster~\cite{sun2025sparse} has been proposed for 3D scene reconstruction, which represents a 3D scene using more efficient sparse voxels and has achieved compelling performance.
Our approach employs sparse voxels as primitives to represent a 3D scene, but unlike SVRaster, we integrate an appearance field, a density field, a feature field, and a confidence field into sparse voxels to comprehensively represent a 3D scene.
Moreover, our approach focuses on exploring the synergy among scene appearance, semantics, and geometry for holistic scene reconstruction and understanding rather than solely on 3D reconstruction.

\subsection{Open-Vocabulary Scene Understanding}
3D open-vocabulary scene understanding focuses on learning 3D scene representations to enable language-guided object recognition and localization beyond a fixed label set~\cite{qin2024langsplat,li2025langsplatv2,peng2026gags,liu2023weakly,kobayashi2022decomposing}.
Although there have been some 2D open-vocabulary scene understanding methods~\cite{li2022language,xu2023open,liang2023open}, they often perform poorly for 3D scene understanding.
To address this problem, some works integrate visual features into point clouds~\cite{zhang2022pointclip,xue2023ulip,lu2023open} to optimize 3D scene representations, while others reconstruct 3D scenes from multi-view images and distill language feature knowledge into 3D scene representations~\cite{shi2024language,qin2024langsplat,li2025vaf,zhucos3d,zhai2025panogs,peng2026gags,zuo2025fmgs,li2025langsplatv2}.
Our work belongs to the latter category.
Our approach differs from existing works in that we distill feature knowledge from both 2D foundation models and geometry foundation models into language and geometry grounded sparse voxel representations to comprehensively represent a 3D scene, facilitating the synergy among scene appearance, semantics, and geometry.

\section{Methodology}
\label{sec:methodology}

\paragraph{\textbf{Problem Statement.}}
This work focuses on holistic scene understanding.
Specifically, given a set of multi-view images capturing a scene, we aim to reconstruct the 3D structure of the scene and embed language grounded semantic features into the 3D scene for open-vocabulary scene understanding.
During training, a 3D scene model is optimized from scratch, where prior feature knowledge is extracted from foundation models and distilled into the 3D scene model to facilitate 3D scene representation learning.
During inference, the optimized 3D scene model can be used to render any view of the 3D scene with high-dimensional feature representations, thereby supporting high-fidelity scene reconstruction and open-vocabulary language queries in 3D space.

\subsection{Preliminary: Sparse Voxel Rasterization}
SVRaster~\cite{sun2025sparse} represents a 3D scene with a sparse voxel grid, which consists of a collection of sparse voxels organized in an Octree structure.
Each voxel is identified by its index $v_i=[i,j,k]$ at an Octree level $l{\in}[1,L]$, so the voxel center $v_c$ and the voxel size $v_s$ are defined as:
\begin{equation}
    \label{eq:voxel}
    v_c=w_c-0.5{\cdot}w_s+v_s{\cdot}v_i,{\quad}v_s=w_s{\cdot}2^{-l},
\end{equation}
where $w_c$ is the Octree center and $w_s$ is the Octree size.
For geometry modeling, SVRaster employs a trilinear density field, which uses eight learnable parameters $v_{geo}{\in}{\mathbb{R}}^{2{\times}2{\times}2}$ corresponding to the corners of each voxel.
For appearance modeling, its appearance field is composed of Spherical Harmonics (SH) $v_{sh}{\in}{\mathbb{R}}^{(N_d+1)^2{\times}3}$, where $N_d$ is the degree of SH.
With $\alpha$-blending and a differentiable rasterizer, the color $C$ of each pixel is computed as:
\begin{equation}
\label{eq:rendering}
C = \sum_{i{\in}N}c_i\alpha_i\prod_{j=1}^{i-1}(1-\alpha_j),
\end{equation}
where $\alpha_i$ and $c_i$ are the alpha and view-dependent color derived from the SH of the $i$th sampled point on the intersected voxel.

\paragraph{\textbf{Limitation.}}
Despite its promising performance, the vanilla SVRaster primarily focuses on 3D reconstruction, neglecting language (semantic) feature field learning for scene understanding.
Although it supports post-processing to lift 2D features onto 3D voxels, this is non-differentiable and does not fully explore the synergy of appearance, semantics, and geometry for holistic scene understanding and reconstruction.

\begin{figure*}[t]
  \centering
   \includegraphics[width=0.9\linewidth]{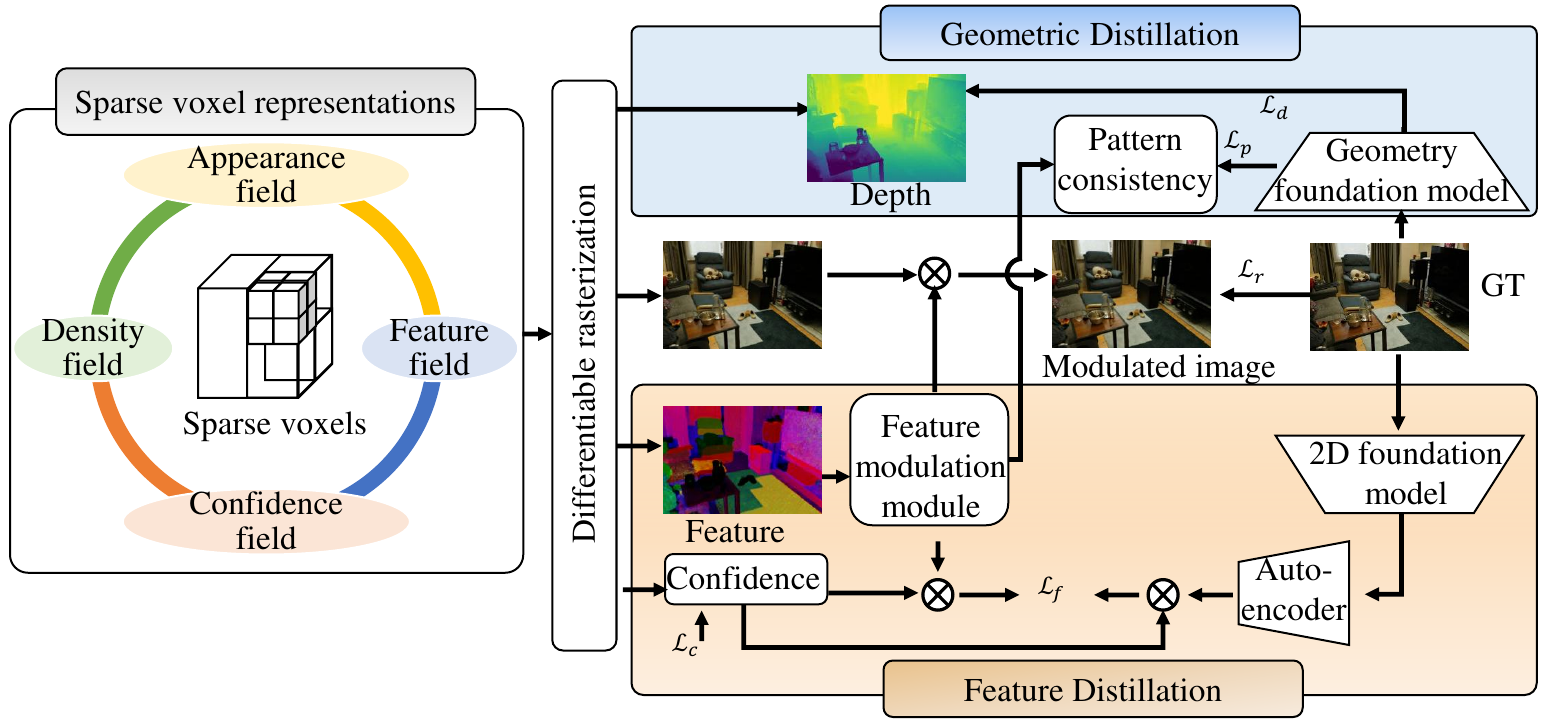}
   \caption{An overview of the proposed approach.
   Our approach optimizes language and geometry grounded sparse voxels as the 3D scene representations to comprehensively model the appearance, semantics, and geometry of the 3D scene in a unified framework.}
   \label{fig_framework}
\end{figure*}

\subsection{The Proposed Approach}

\paragraph{\textbf{Sparse Voxel Representations.}}
An overview of the proposed LangSVR approach is depicted in Fig.~\ref{fig_framework}.
We employ an appearance field, a density field, a feature field, and a confidence field to comprehensively represent a 3D scene.
Among them, the appearance field and the density field are similar to those of SVRaster, while the feature field and the confidence field are introduced to optimize 3D sparse voxel feature representations for scene understanding, and they are differentiable.
For each voxel, we employ a learnable parameter $v_f$ to learn the corresponding feature embedding and a learnable parameter $v_{conf}$ to learn the confidence of each voxel.
Since we also employ $\alpha$-blending and a differentiable rasterizer for rendering, a feature map $m_r$ and a confidence map $m_c$ can be obtained with Eq.~\eqref{eq:rendering} by replacing $c_i$ with $v_f$ or $v_{conf}$.

\begin{figure}[t!]
    \centering
     \includegraphics[width=0.99\linewidth]{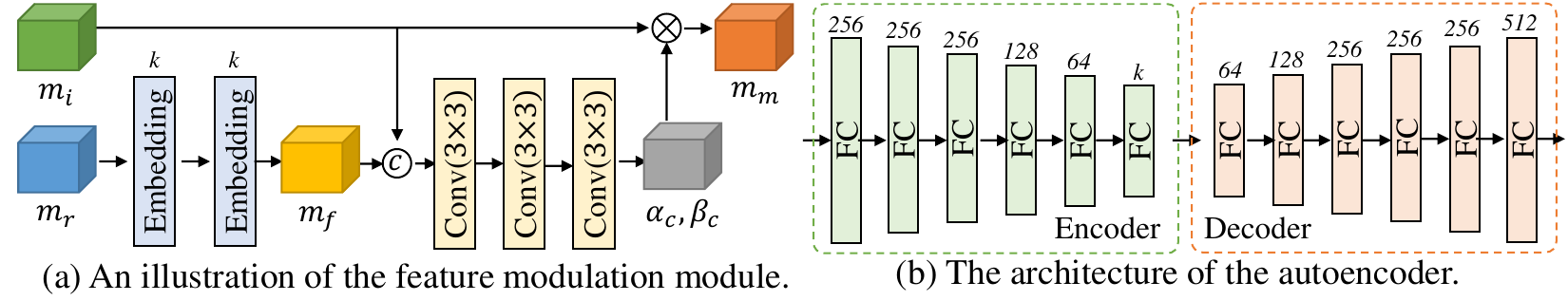}
     \caption{
        The illustrations of (a) the proposed feature modulation module and (b) the architecture of the employed autoencoder.
        We show the hidden dimension of each layer above the corresponding layer.
        }
     \label{fig_module_architecture}
\end{figure}

\paragraph{\textbf{Feature Modulation.}}
To distill language features from a 2D foundation model into $v_f$, we extract a feature map for each view using a frozen foundation model and minimize the difference between the rendered feature map and the extracted feature map to optimize 3D scene representations.
However, it is impractical to directly optimize a high-dimensional feature (\eg, 512-dim in CLIP) for $v_f$, which is computationally intensive.
Hence, similar to~\cite{qin2024langsplat}, we pre-train an autoencoder (see Fig.~\ref{fig_module_architecture}(b)) to map features $m_{a}$ into a $k$-dimensional latent space, where $k{\ll}512$, and define the voxel feature embedding as $v_f{\in}{\mathbb{R}^{1{\times}k}}$.
Next, we construct a feature modulation module (see Fig.~\ref{fig_module_architecture}(a)) to promote the synergy of the appearance field, the density field, and the feature field.
Specifically, we employ two embedding layers ($f_{e1}$ and $f_{e2}$) to modulate the rendered features and obtain the modulated features $m_f$.
This is defined as:
\begin{equation}
    \label{eq:projection}
    m_f=F_s(m_r{\cdot}f_{e1}){\cdot}f_{e2},
\end{equation}
where $F_s$ is a softmax function.
Eq.~\eqref{eq:projection} helps to project the rendered features into a compact latent space by aggregating a fixed set of embeddings with learned weights, which facilitates the optimization of language features.
Then, inspired by~\cite{wu2025armgs}, we use a lightweight module $F_{c}$ with three convolutional layers to further process the modulated feature, which modulates the rendered image $m_i$ to obtain the modulated image $m_m$.
This is defined as:
\begin{align}    
    \{\alpha_c,\beta_c\}&=F_{c}(Concat(m_f,m_i)), \label{eq:appearance_param}\\
    m_m&={\alpha_c}{\cdot}m_i+{\beta_c}. \label{eq:appearance}    
\end{align}
Here, $\{\alpha_c,\beta_c\}$ are per-pixel modulation parameters, and $Concat$ is the concatenation operation.
Eqs.~\eqref{eq:appearance_param} and~\eqref{eq:appearance} help to modulate scene appearance based on learned language grounded semantic features, which enhances the synergy between appearance and semantics for holistic scene representation learning.

\paragraph{\textbf{Confidence Regularization.}}
To transfer the knowledge of language feature from the 2D foundation model into our 3D scene model, we employ a feature distillation loss $\mathcal{L}_f$ to minimize the difference between the modulated feature map $m_f$ and the extracted language feature map $m_{a}$.
However, it is inevitable that feature maps extracted from a 2D foundation model may be noisy and inconsistent across views.
To alleviate this problem, we employ a confidence field to generate a confidence map $m_c$ for each view and use it to filter out noisy representations during feature distillation.
Hence, $\mathcal{L}_f$ is defined as:
\begin{equation}
    \label{eq:feature_distance}    
    \mathcal{L}_f=||F_g(m_c){\cdot}m_f, F_g(m_c){\cdot}m_{a}||_1,
\end{equation}
where $F_g$ is a sigmoid function.
To avoid a trivial solution when using $m_c$ in Eq.~\eqref{eq:feature_distance}, we use a confidence map regularization ${\mathcal{L}_c}$ as:
\begin{equation}
    \label{eq:confidence}    
    \mathcal{L}_c=\mathbb{E}[-\log(F_g(m_c))],
\end{equation}
where $\mathbb{E}$ denotes the mean operation over all pixels.

\paragraph{\textbf{Geometric Distillation.}}
In addition to appearance and semantic modeling, capturing the underlying geometric structure of 3D scenes is also of great importance for holistic scene understanding and reconstruction. 
Although sparse voxel rasterization has shown better scene geometry quality than 3DGS, we observe that its depth estimation can sometimes be noisy, especially in high-frequency areas.
To enhance scene geometric constraints, we distill prior depth knowledge from a geometry foundation model into our 3D scene representations via a depth correlation regularization~\cite{xiong2025sparsegs}.
This is defined as:
\begin{equation}
    \label{eq:depth}    
    \mathcal{L}_d=\mathbb{E}[1-(\frac{D_r-\mathbb{E}[D_r]}{std(D_r-\mathbb{E}[D_r])})(\frac{D_d-\mathbb{E}[D_d]}{std(D_d-\mathbb{E}[D_d])})],
\end{equation}
where $D_r$ and $D_d$ are the rendered depth and the prior depth, respectively, and $std$ is the standard deviation.
Meanwhile, we propose transferring geometry grounded feature $m_{g}$ knowledge from the geometry foundation model to 3D scene representation to further regularize the optimization of the modulated features.
Although the modulated features and the geometry features may have different distributions, which makes it difficult to directly minimize their difference, we consider that their local patterns should be consistent.
In light of this, we employ a pattern consistency regularization to align the local patterns between the modulated features $m_f$ and the geometry grounded features $m_{g}$, which is defined as:
\begin{equation}
    \label{eq:pattern}    
    \mathcal{L}_p=||F_n(m_f), F_n(m_{g})||_1,
\end{equation}
where $F_n$ is a function for calculating similarity between neighboring embeddings, and the dimensions of its inputs do not need to be the same.

\paragraph{\textbf{Model Optimization.}}
Our approach can be trained from scratch for both 3D scene reconstruction and feature field distillation in a unified framework.
We define the overall training objective as follows:
\begin{equation}
    \label{eq:loss}    
    \mathcal{L}=\mathcal{L}_r+\lambda_1\mathcal{L}_f+\mathcal{L}_c+\mathcal{L}_p+\lambda_2\mathcal{L}_d,
\end{equation}
where $\lambda_1$ and $\lambda_2$ are weight coefficients, and $\mathcal{L}_r$ is the image reconstruction loss following~\cite{sun2025sparse}.

\section{Experiment}
\label{sec:experiment}

\subsection{Datasets and Evaluation Protocol}
\paragraph{\textbf{Datasets.}}
We conduct experiments on the LERF dataset~\cite{kerr2023lerf} and the Mip-NeRF360 dataset~\cite{barron2022mip}.
On the LERF dataset, we use four scenes, \ie, Ramen, Teatime, Kitchen, and Figurines, following~\cite{qin2024langsplat}.
On the Mip-NeRF360 dataset, we use four scenes, \ie, Room, Counter, Bonsai, and Garden, following~\cite{peng2026gags}.
We perform evaluation for open-vocabulary scene understanding and 3D scene reconstruction on these datasets.
Specifically, for scene understanding, we follow~\cite{peng2026gags,li2025langsplatv2,qin2024langsplat} to evaluate 3D semantic segmentation and 3D object localization tasks.
On these tasks, the testing ground-truth annotations are provided by~\cite{qin2024langsplat} on the LERF dataset and by~\cite{peng2026gags} on the Mip-NeRF360 dataset.
For scene reconstruction, we follow~\cite{sun2025sparse,kerbl20233d} and evaluate the novel view synthesis task using every eighth frame as a testing view. 

\paragraph{\textbf{Evaluation Metrics.}}
For scene understanding, we follow the evaluation setup of~\cite{li2025langsplatv2,qin2024langsplat}.
We use the mean Intersection over Union (mIoU$\uparrow$) to evaluate 3D semantic segmentation and the mean Accuracy (mAcc$\uparrow$) to evaluate 3D object localization.
For scene reconstruction, we use the novel view synthesis setting and employ PSNR$\uparrow$ and LPIPS$\downarrow$ as metrics, following~\cite{sun2025sparse,kerbl20233d}.

\paragraph{\textbf{Implementation Details.}}
We implement our approach with Python and PyTorch.
We use SVRaster~\cite{sun2025sparse} as the baseline model to build our approach.
For the 2D foundation model, following~\cite{qin2024langsplat}, we use CLIP~\cite{radford2021learning} and SAM~\cite{kirillov2023segment} to pre-process the datasets for extracting language features.
For the geometry foundation model, such as VGGT~\cite{wang2025vggt} and Depth-Anything-V2~\cite{yang2024depth}, we pre-process the datasets for extracting geometry grounded priors and embeddings.
We empirically set $k=32$, $\lambda_1=0.1$, and $\lambda_2=0.01$.
We train our 3D model with the Adam~\cite{kingma2014adam} optimizer for 20000 iterations.
Besides, we construct the autoencoder~\cite{peng2026gags,qin2024langsplat} by using six linear layers for both the encoder and the decoder, respectively, with ReLU being used between layers. 
The hidden dimension of the autoencoder is shown in Fig.~\ref{fig_module_architecture}(b).
The autoencoder is pre-trained with the Adam optimizer for 200 epochs.

\paragraph{\textbf{Application.}}
Our model supports various tasks, such as 3D semantic segmentation, 3D object localization, novel view synthesis, depth and normal map rendering, and mesh extraction.
This enables holistic scene understanding and reconstruction.
For novel view synthesis and depth/normal map rendering, our approach can directly render images from a given camera viewpoint.
For open-vocabulary 3D semantic segmentation and 3D object localization, we decode high-dimensional features using the learned feature embeddings and the pre-trained autoencoder following~\cite{qin2024langsplat}, and calculate the similarity between the embedded language query and the decoded feature embeddings for semantic segmentation and object localization.
For mesh extraction, our approach inherits the capabilities of SVRaster and can directly extract meshes following~\cite{sun2025sparse}.

\begin{table*}[t!]
    \caption{Quantitative comparison with the state-of-the-arts on the LERF and Mip-NeRF360 datasets.
    We report average results across four scenes for each dataset.
    $^\star$: 3D reconstruction results for these two-stage methods are borrowed from 3DGS, as they use pre-trained 3DGS for feature distillation.
    $^\dagger$: Lifting fine-grained features to 3D primitives.
    Some results are cited from existing work.
    }       
    \centering    
    \begin{tabular}{lcccccccc}
        \hline
        \multirow{3}{*}{Method} & \multicolumn{4}{c}{LERF dataset} & \multicolumn{4}{c}{Mip-NeRF360 dataset}\\
        \cmidrule(lr){2-5} \cmidrule(lr){6-9}
        & \multicolumn{2}{c}{Understanding} & \multicolumn{2}{c}{Reconstruction} & \multicolumn{2}{c}{Understanding} & \multicolumn{2}{c}{Reconstruction}\\
        \cmidrule(lr){2-3} \cmidrule(lr){4-5} \cmidrule(lr){6-7} \cmidrule(lr){8-9}
        & mIoU$\uparrow$ & mAcc$\uparrow$ & PSNR$\uparrow$ & LPIPS$\downarrow$ & mIoU$\uparrow$ & mAcc$\uparrow$ & PSNR$\uparrow$& LPIPS$\downarrow$ \\
        \midrule
        \midrule
        LERF\cite{kerr2023lerf}              &37.4  & 73.6 & 20.75   & 0.264     &- &-  &-  &-  \\
        Feature-3DGS\cite{zhou2024feature}   &38.3  & 70.1 & \cellcolor{tabsecond}{23.98} & 0.222     &- &-  &-  &-      \\  
        LangSplat$^\star$\cite{qin2024langsplat}    &51.4  & \cellcolor{tabsecond}{84.3} &{23.89} &{0.218}  & 54.7  & {73.1}  & {29.38}& {0.183} \\
        LEGaussians\cite{shi2024language}    &24.5  & 67.4 & -  & -     & 29.1  & 65.2  & -  & -     \\
        GS-Grouping\cite{ye2024gaussian}     & 36.6 & 49.1 & 23.20  & 0.256     & 49.1  & 65.0     & 28.22     & \cellcolor{tabthird}{0.179}    \\
        GOI$^\star$\cite{qu2024goi}                 &42.0  & 59.2 &{23.89} &{0.218} & 58.5  & 68.9  & {29.38}& {0.183} \\
        VaF-LangSplat$^\star$\cite{li2025vaf}       &\cellcolor{tabthird}{56.5}  & 74.2 &{23.89} &{0.218} &- &-  &-  &-  \\
        CF$^3$\cite{lee2025cf3}              &54.0  & 83.7 & 23.84 & 0.222  & 59.2   &\cellcolor{tabthird}{83.9}  &27.02 &0.239  \\
        LangSplatV2$^\star$\cite{li2025langsplatv2}  &\cellcolor{tabsecond}{59.9}  & \cellcolor{tabthird}{84.1} &{23.89} &{0.218} & \cellcolor{tabsecond}{69.4} & 81.8  & {29.38} & {0.183} \\
        GAGS$^\star$\cite{peng2026gags}              & 54.1  & {81.7} &{23.89} &{0.218} & \cellcolor{tabthird}{64.5}  & \cellcolor{tabsecond}{88.7} & {29.38}& {0.183} \\
        \midrule
        3DGS$^\dagger$\cite{kerbl20233d}             &43.5   & 71.9   & \cellcolor{tabthird}{23.89}    & \cellcolor{tabthird}{0.218}            & 47.8       & 79.5  & \cellcolor{tabthird}{29.38}     & {0.183}         \\                
        SVRaster$^\dagger$\cite{sun2025sparse}       &46.1 & 61.1   & {23.88} & \cellcolor{tabsecond}{0.215} & 49.7 & 70.0  & \cellcolor{tabsecond}{29.66} &\cellcolor{tabsecond}{0.160} \\
        \midrule
        LangSVR (Ours)                           &\cellcolor{tabfirst}{62.1}  & \cellcolor{tabfirst}{84.4} & \cellcolor{tabfirst}{24.02} & \cellcolor{tabfirst}{0.212} & \cellcolor{tabfirst}{71.2}  & \cellcolor{tabfirst}{89.4}  & \cellcolor{tabfirst}{29.87} & \cellcolor{tabfirst}{0.159} \\
        \hline
    \end{tabular}
    \label{tab:exp-sota-lerf-mipnerf360}
\end{table*}

\subsection{Comparison with the State-of-the-Arts}
\paragraph{\textbf{Compared Methods.}}
Since the main focus of this work is learning effective 3D scene representations to facilitate holistic scene understanding and reconstruction, we mainly compare with state-of-the-art feature field distillation methods, including LERF~\cite{kerr2023lerf}, Feature-3DGS~\cite{zhou2024feature}, LangSplat~\cite{qin2024langsplat}, LEGaussians~\cite{shi2024language}, GS-Grouping~\cite{ye2024gaussian}, GOI~\cite{qu2024goi}, VaF-LangSplat~\cite{li2025vaf}, CF$^3$~\cite{lee2025cf3}, LangSplatV2 (LS-V2)~\cite{li2025langsplatv2}, and GAGS~\cite{peng2026gags}.
In addition, we also compare with two state-of-the-art 3D scene reconstruction methods, namely 3DGS~\cite{kerbl20233d} and SVRaster~\cite{sun2025sparse}.

\begin{figure*}[t!]
  \centering
   \includegraphics[width=0.85\linewidth]{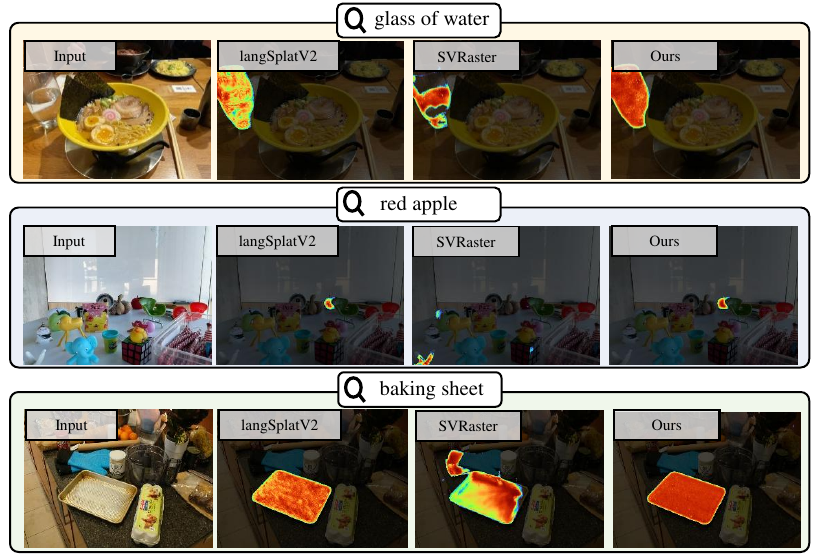}
   \caption{Qualitative comparison with state-of-the-art methods in 3D semantic segmentation on the LERF and Mip-NeRF360 datasets.
   The segmented regions are highlighted.}
   \label{fig_sota_segmentation}
\end{figure*}

\begin{figure*}[t!]
  \centering
   \includegraphics[width=0.85\linewidth]{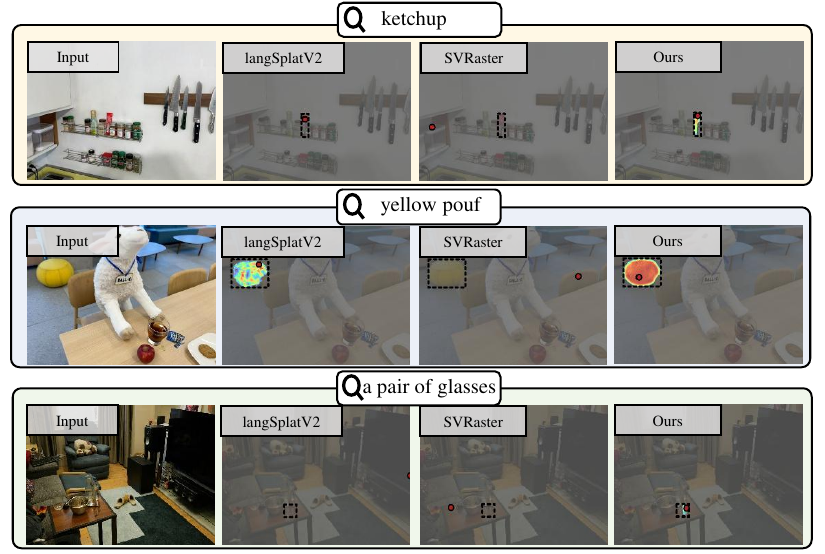}
   \caption{Qualitative comparison with state-of-the-art methods in 3D object localization on the LERF and Mip-NeRF360 datasets.
   Red dots denote the positions with the highest localization responses and black dashed boxes represent the ground-truth localizations.}
   \label{fig_sota_localization}
\end{figure*}

\begin{figure*}[t!]
  \centering
   \includegraphics[width=0.85\linewidth]{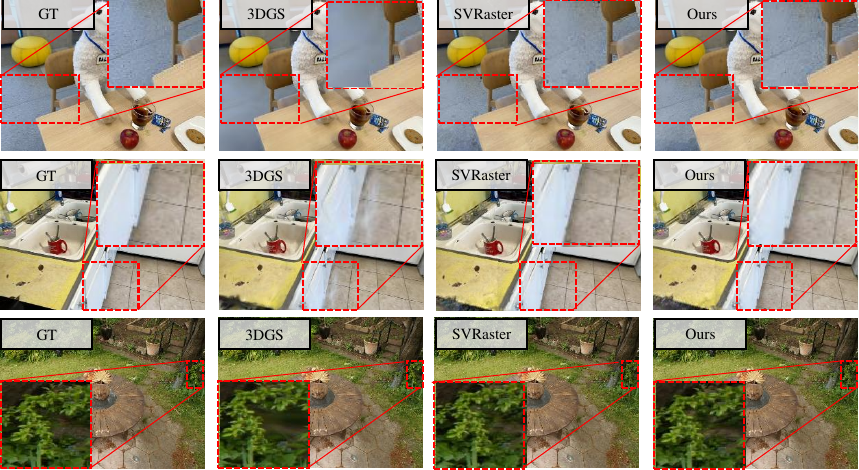}
   \caption{Qualitative comparison with state-of-the-art methods in novel view synthesis on the LERF and Mip-NeRF360 datasets.
   Red dashed boxes highlight some fine-grained details.}
   \label{fig_sota_reconstruction}
\end{figure*}

\paragraph{\textbf{Quantitative Comparison and Result Analysis.}}
We report quantitative results for open-vocabulary scene understanding and 3D scene reconstruction in Tab.~\ref{tab:exp-sota-lerf-mipnerf360}.
Our approach (LangSVR) achieves superior overall performance compared with state-of-the-art methods.
Specifically, on the LERF dataset, for scene understanding, our approach achieves an mIoU of 62.1 in 3D semantic segmentation and an mAcc of 84.4\% in 3D object localization, outperforming state-of-the-art methods such as GAGS and LangSplatV2.
For scene reconstruction, our approach achieves 24.02 dB in PSNR and 0.212 in LPIPS, surpassing 3DGS and SVRaster.
On the Mip-NeRF360 dataset, our approach improves the state-of-the-art by 1.8 in mIoU for 3D semantic segmentation and by 0.7\% in mAcc for 3D object localization.
For scene reconstruction, our approach achieves the best LPIPS (0.159) and PSNR (29.87 dB), demonstrating competitive performance compared with state-of-the-art methods.
These improvements can be attributed to the synergy of the proposed components for learning language and geometry grounded sparse voxel representations in a unified framework.
Furthermore, as illustrated in Fig.~\ref{fig_illustration}, our approach achieves better overall performance (located in the top right corner) compared with the state-of-the-arts for holistic scene understanding and reconstruction.

\paragraph{\textbf{Qualitative Comparison and Result Analysis.}}
We present qualitative comparisons for 3D semantic segmentation in Fig.~\ref{fig_sota_segmentation}, 3D object localization in Fig.~\ref{fig_sota_localization}, and 3D scene reconstruction in Fig.~\ref{fig_sota_reconstruction}.
From these results, we can observe that for scene understanding, our approach produces more accurate results in both semantic segmentation and object localization given language queries.
For example, given the query ``glass of water'' in 3D semantic segmentation, our approach yields an accurate segmentation, whereas SVRaster produces an incomplete segmentation and LangSplatV2 generates a result with low response values.
Similarly, for the query ``a pair of glasses'' in 3D object localization, our approach accurately locates the glasses, while other methods fail to identify the correct region.
For scene reconstruction, our approach achieves high-fidelity results with finer details, such as the texture of the floor and the surface of the cabinet in Fig.~\ref{fig_sota_reconstruction}.
These results demonstrate the superiority of our approach over state-of-the-art methods.

\begin{table*}[t!]
    \caption{Effectiveness and efficiency against SVRaster on the LERF dataset.    
    $^\dagger$Results are obtained with a V100 GPU.
    $^\star$Training GPU memory in terms of GB.
    }
    \resizebox{0.99\columnwidth}{!}{
    \centering    
    \begin{tabular}{l|cc|cc|c|c}
        \hline
        \multirow{2}{*}{Method} & \multicolumn{2}{c|}{Scene Understanding} & \multicolumn{2}{c|}{Scene Reconstruction} & {Speed$^\dagger$} & {GPU}\\
        & Seg.(mIoU$\uparrow$) & Loc.(mAcc$\uparrow$) & PSNR$\uparrow$ & LPIPS$\downarrow$ & (Image FPS$\uparrow$) & Memory$\downarrow$$^\star$\\
        \hline
        \hline
        SVRaster & 46.1 & 61.1 & 23.88 & 0.215 & \cellcolor{tabfirst}{69}  &   \cellcolor{tabfirst}{12}  \\
        Ours     & \cellcolor{tabfirst}{62.1}  & \cellcolor{tabfirst}{84.4} & \cellcolor{tabfirst}{24.02} & \cellcolor{tabfirst}{0.212} & 35  &  14 \\
        \hline
    \end{tabular}
    }      
    \label{tab:exp-ours-svraster}
\end{table*}

\begin{figure*}[t!]
  \centering
   \includegraphics[width=0.85\linewidth]{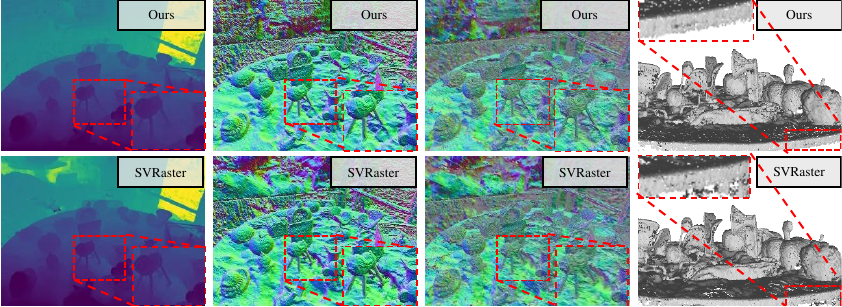}
   \caption{Visualizations of the rendered depth maps, depth to normal maps, normal maps, and extracted meshes on the LERF dataset.}
   \label{fig_depth_mesh}
\end{figure*}

\subsection{Ablation Study}

\paragraph{\textbf{Effectiveness and Efficiency Compared to the SVRaster Baseline.}}
Since our approach is built on SVRaster, we compare the effectiveness and efficiency in Tab.~\ref{tab:exp-ours-svraster}.
We can see that for scene understanding, our approach significantly improves the baseline SVRaster by relatively 34.7\% in 3D semantic segmentation and by relatively 38.1\% in 3D object localization.
For 3D scene reconstruction, our approach improves the baseline by relatively 1.4\% in LPIPS and by relatively 0.6\% in PSNR for novel view synthesis.
For efficiency evaluation, we measure the speed (FPS) for image rendering and report the training GPU memory (GB) consumption.
Due to the additional components designed to promote the synergy of appearance, semantics, and geometry, our approach renders images more slowly than SVRaster and requires slightly more GPU memory.
Additionally, we visualize the rendered depth maps, depth-to-normal maps, normal maps, and extracted meshes in Fig.~\ref{fig_depth_mesh}.
It can be seen that our approach models scene geometry better than SVRaster.
These results verify that our approach achieves superior holistic scene understanding and reconstruction compared with the baseline SVRaster, representing a better trade-off between accuracy and speed.

\begin{figure*}[t]
  \centering
   \includegraphics[width=0.85\linewidth]{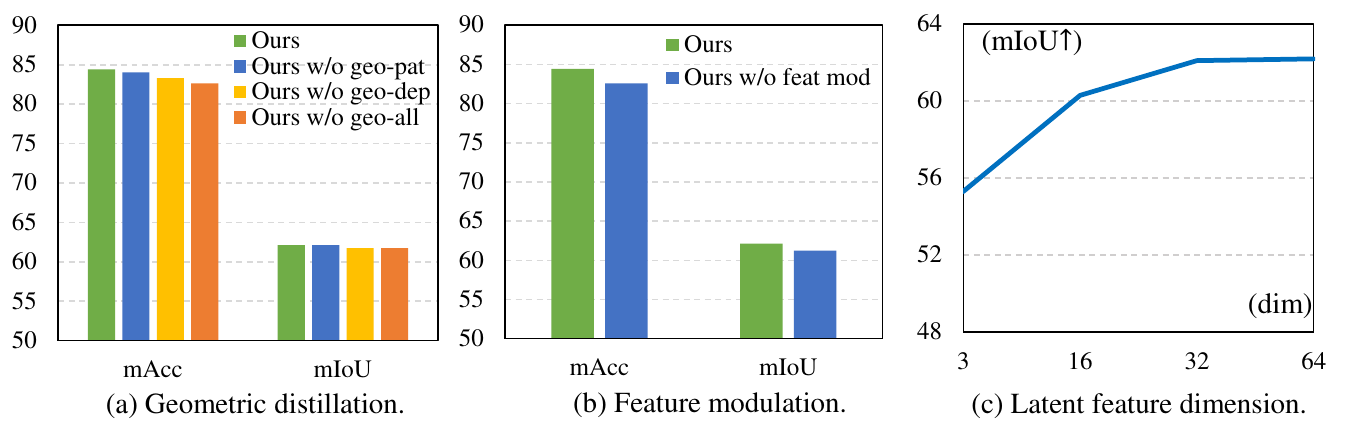}
   \caption{Evaluation of the effectiveness of (a) geometric distillation, (b) feature modulation, and (c) latent feature dimension $k$ in the proposed method.
   Results on the LERF dataset are reported.}
   \label{fig_exp_geo_feat}
\end{figure*}

\paragraph{\textbf{Analysis of Geometric Distillation.}}
For the geometry foundation model, we evaluate both VGGT~\cite{wang2025vggt} and Depth-Anything-V2~\cite{yang2024depth} and report the results in Tab.~\ref{tab:exp-geo-model}.
Overall, both variants achieve comparable performance.
Ours with VGGT yields slightly better PSNR for 3D scene reconstruction, while ours with Depth-Anything-V2 achieves slightly better mIoU for 3D semantic segmentation.
Furthermore, as shown in Fig.~\ref{fig_exp_geo_feat}(a), when removing the pattern consistency regularization or the depth correlation regularization, the performance of our approach degrades, while removing both regularizations leads to a more significant drop.
In addition, as shown in Fig.~\ref{fig_depth_mesh}, our approach with geometric distillation also outperforms the baseline SVRaster for geometric reconstruction.
These results demonstrate that the proposed geometric distillation contributes positively to the overall performance.

\begin{table*}[t!]
    \caption{Evaluation of using different geometric foundation models in the proposed approach.
    Results on the LERF dataset are reported.
    }
    \centering    
    \begin{tabular}{l|cc|cc}
        \hline
        \multirow{2}{*}{Variant} & \multicolumn{2}{c|}{Understanding} & \multicolumn{2}{c}{Reconstruction} \\
        & mIoU$\uparrow$ & mAcc$\uparrow$ & PSNR$\uparrow$ & LPIPS$\downarrow$ \\
        \hline
        \hline
        
        Ours w/ VGGT~\cite{wang2025vggt}     & 62.0  & 84.4 &   24.06&   0.212   \\
        Ours w/ Depth-Anything-V2~\cite{yang2024depth}      & 62.1  & 84.4 &   24.02&   0.212   \\
        \hline
    \end{tabular}
    \label{tab:exp-geo-model}
\end{table*}

\begin{wraptable}{r}{0.45\textwidth}
    \vspace{-1.0cm}
    \caption{Analysis of the confidence field (results on the LERF dataset).
    }
    \centering    
    \begin{tabular}{l|cc}
        \hline        
        Variant         & mAcc$\uparrow$ & PSNR$\uparrow$ \\
        \hline
        \hline        
        Ours            & 84.4           & 24.02\\
        Ours w/o conf.  & 82.2           & 24.07\\
        \hline
    \end{tabular}     
    \label{tab:exp-conf}
    \vspace{-0.5cm}
\end{wraptable}
\paragraph{\textbf{Analysis of Feature Distillation.}}
We analyze the effect of the proposed feature modulation module in Fig.~\ref{fig_exp_geo_feat}(b).
Our approach without the feature modulation module yields significantly worse performance.
For example, in terms of mIoU, ours with feature modulation achieves 62.1, while ours without feature modulation achieves 61.2.
Furthermore, we evaluate the impact of different latent space dimensions $k$ in Fig.~\ref{fig_exp_geo_feat}(c).
We can see that when $k=3$, the performance is worse, indicating that highly compressed features are insufficient for modeling representations in the latent space.
Increasing the feature dimension from $3$ to $32$ leads to a substantial improvement, while setting $k=64$ also brings gains.
Therefore, we set $k=32$ to achieve a better balance between effectiveness and efficiency.
Additionally, we show the effect of the confidence field in Tab.~\ref{tab:exp-conf}.
Removing the confidence field from our approach results in degraded performance for 3D semantic segmentation but yields comparable results for 3D reconstruction.
This suggests that confidence maps are helpful for resolving multi-view inconsistency in scene semantic representation learning, but are less critical for 3D reconstruction.

\begin{figure}[t!]
  \centering
   \includegraphics[width=0.85\linewidth]{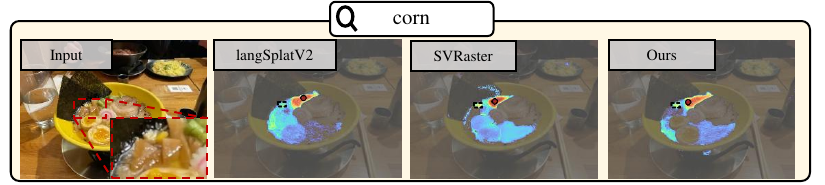}
   \caption{An example of a failure case.
   All evaluated methods failed to locate the corn in the bowl since the corn in the bowl is very small.
   }
   \label{fig_exp_failure_case}
\end{figure}

\subsection{Limitation and Future Work}
Although our approach achieves compelling performance for holistic scene understanding and reconstruction, it still has some limitations.
For instance, our approach may not model all fine-grained details of 3D scenes.
As shown in Fig.~\ref{fig_exp_failure_case}, all evaluated methods, including our LangSVR approach, failed to locate the corn in the bowl.
This case is particularly challenging due to the very small size of the corn kernel and the complex background.
Additionally, since it is impractical to directly optimize high-dimensional features for a large number of primitives, we follow existing work~\cite{qin2024langsplat} and use a pre-trained autoencoder to map language features into a low-dimensional latent space.
However, the performance of our model may be limited by the autoencoder, and inadequate representation of language features may lead to suboptimal performance, as discussed in the experiment.
Our future work aims to resolve these limitations and further explore our approach for more challenging spatial understanding and reasoning tasks.

\section{Conclusion}
\label{sec:conclusion}
In this work, we introduce a novel approach named LangSVR for holistic scene understanding and reconstruction.
Our approach learns language and geometry grounded sparse voxel representations to comprehensively model scene appearance, semantics, and geometry in a unified framework, which synergistically integrates feature field distillation and geometric distillation.
Extensive experiments demonstrate that our approach achieves competitive performance compared with state-of-the-art methods.
Our in-depth ablation study further verifies the effectiveness of the proposed components.

\bibliographystyle{plain}
\bibliography{ref}

\end{document}